\documentclass[10pt, a4paper]{article}

\usepackage[final]{lrec2026}
\usepackage{booktabs}
\usepackage{tabularx}
\usepackage{wrapfig}

\DeclareUnicodeCharacter{202F}{~}
\DeclareUnicodeCharacter{2011}{-}

\title{CareMedEval dataset: Evaluating Critical Appraisal and Reasoning in the Biomedical Field}

\name{
Doria Bonzi\textsuperscript{1},
Alexandre Guiggi\textsuperscript{2},
Frédéric Béchet\textsuperscript{3},
Carlos Ramisch\textsuperscript{3},
Benoit Favre\textsuperscript{{3},{4}}
}

\address{
\textsuperscript{1}University of Lorraine, LORIA, France \\
\textsuperscript{2}University Grenoble-Alpes, France \\
\textsuperscript{3}Aix-Marseille University, LIS, France \\
\textsuperscript{4}CNRS, Grenoble INP, LIG, France\\
doria.bonzi@loria.fr, alexandre.guiggi@gmail.com, \{frederic.bechet, carlos.ramisch, benoit.favre\}@lis-lab.fr
}

 \abstract{
Critical appraisal of scientific literature is an essential skill in the biomedical field. While large language models (LLMs) can offer promising support in this task, their reliability remains limited, particularly for critical reasoning in specialized domains. We introduce CareMedEval, an original dataset designed to evaluate LLMs on biomedical critical appraisal and reasoning tasks. Derived from authentic exams taken by French medical students, the dataset contains 534 questions based on 37 scientific articles. Unlike existing benchmarks, CareMedEval explicitly evaluates critical reading and reasoning grounded in scientific papers. Benchmarking state-of-the-art generalist and biomedical-specialized LLMs under various context conditions reveals the difficulty of the task: open and commercial models fail to exceed an Exact Match Rate of 0.5 even though generating intermediate reasoning tokens considerably improves the results. Yet, models remain challenged especially on questions about study limitations and statistical analysis. CareMedEval provides a challenging benchmark for grounded reasoning, exposing current LLM limitations and paving the way for future development of automated support for critical appraisal.
\\ \newline  }

\begin{document}
\maketitleabstract

\section{Introduction}
Medical professionals must engage in continuous learning to stay up to date with evolving medical knowledge. Even though they can gather knowledge from trusted sources such as Cochrane, they also engage with latest research published in scientific papers often available before peer review in dedicated archives.

Critically appraising scientific publications is a complex cognitive task, even for trained physicians. As highlighted by previous studies, proper interpretation of biomedical literature requires not only familiarity with medical content, but also awareness of methodology and statistics~\citep{duprel2009critical}. Moreover, studies have shown that medical research can suffer from major methodological flaws, raising concerns about the reliability and overall trustworthiness of scientific evidence (\citealp{ioannidis2005why}; \citealp{begley2012raise}). These issues often relate to study design quality and various forms of bias (\citealp{chalmers2009avoidable}; \citealp{dickersin1987publication}). These challenges show why teaching and assessing critical appraisal skills remains a major and ongoing issue in the biomedical field.

Natural language processing (NLP), and in particular large language models (LLMs), represents a promising technological solution for supporting this lifelong learning process.
Recent development of LLMs has put focus on their "reasoning capabilities", opening new possibilities for supporting medical professionals in the critical reading, analysis, and synthesis of scientific literature. It is  important to evaluate the reliability of such technologies.

Several evaluation benchmarks already exist in the biomedical domain, but these resources do not explicitly target the evaluation of research methodology or a system's ability to identify limitations and biases in a study. As a result, they are not well-suited for measuring the specific skills involved in critical appraisal.

Despite the growing interest in applying LLMs to these tasks, there remain significant challenges related to hallucinations, bias, and keeping  scientific accuracy intact (\citealp{wang2024surveylargelanguagemodels}; \citealp{yun2023appraisingpotentialusesharms}; \citealp{meng2024llm_medicine}). Recent work has explored long-context processing in LLMs (\citealp{nelson2024needlehaystackmemorybased}; \citealp{li_longcancontextlength_2023}) and their application to the medical field (\citealp{bazoge2024adaptationbiomedicalclinicalpretrained}), including retrieval-augmented generation (RAG) systems, where external knowledge sources are integrated into the model's reasoning process (\citealp{10.1093/jamia/ocaf008}). In these studies, RAG methods make biomedical QA more accurate and robust, yet they mainly support information access and synthesis. The task of critically assessing study design and validity is not well represented in benchmarks, especially when grounded in a given study described in a scientific article.

In this work, we introduce (1) a dataset for evaluating critical appraisal of medical studies described in scientific articles\footnotemark[1], derived from multiple choice question-answers (MCQA) medical education exams in French. With this dataset release, we present (2) a comprehensive evaluation of diverse state-of-the-art language models on this challenging task, providing baseline performance results and insights into model capabilities and limitations.

\footnotetext[1]{Dataset and code available at \url{https://github.com/bonzid/CareMedEval}.}

\begin{figure*}[htbp]
\centering
\footnotesize
\begin{tcolorbox}[
  title=Sample question with context and answer choices taken from the dataset,
  colback=gray!5,
  colframe=gray!50!black,
  fonttitle=\bfseries,
  enhanced jigsaw,
  width=\textwidth,
  boxrule=0.2pt,
  sidebyside
]

\small
\textbf{Article, available in PDF or plain text:} \\

\begin{center}
\includegraphics[width=1\linewidth]{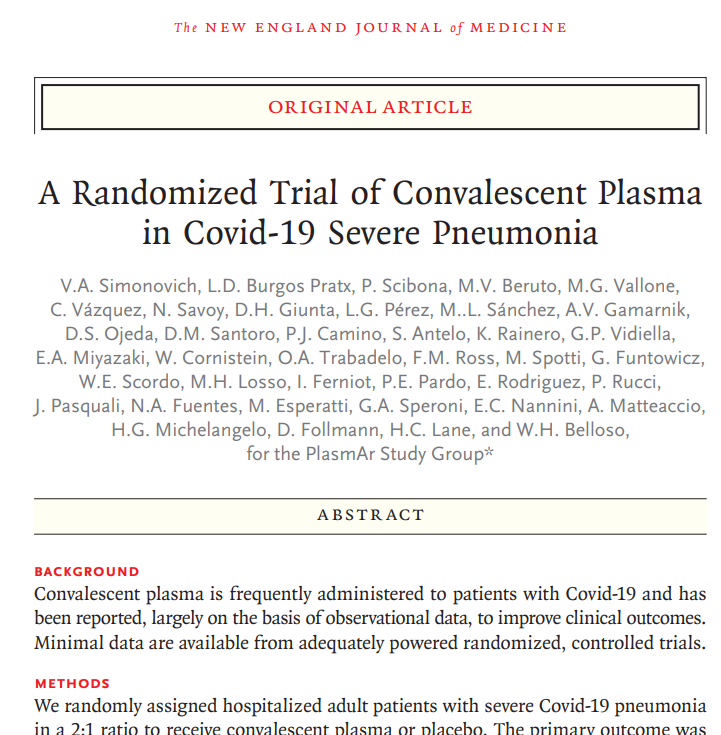}
\end{center}

\tcblower
{\fontsize{1}{2}\selectfont
\small
You are a physician capable of rigorously interpreting data from medical studies. Based on the given article, answer the following multiple-choice question.
Provide only the letter(s) corresponding to the correct answer(s) among: A, B, C, D, E.
Your answer must strictly follow the exact format: one or more letters, separated by commas (e.g., B, E).

\vspace{1em}
\small
\textbf{Question:} \\
\textit{What is the main limitation of this study?}

\vspace{0.5em}
\small
\textbf{Answers:}
\begin{enumerate}[label=\textbf{\alph*)},leftmargin=2em]
    \item The 2:1 randomization
    \item The choice of the primary endpoint
    \item The number of subjects included
    \item The amount of antibodies contained in the plasma
    \item The fact that the study was conducted in Argentina
\end{enumerate}
}
\end{tcolorbox}
\caption{Example from the dataset showing an excerpt from a scientific article, the given instruction prompt with a corresponding question, and answer choices.}
\label{fig:multimodataset}
\end{figure*}

Unlike existing biomedical question answering datasets which are typically not grounded in research articles, our resource is directly contextualized with authentic scientific publications, aiming to evaluate the information gathering and reasoning capabilities of models. 

We argue that this dataset is complementary both to factual and RAG-oriented medical benchmarks, by addressing the underexplored dimension of critical appraisal in the medical field. The dataset developed in this work could also support future technologies beyond LLMs, providing a foundation for the development of tools aimed at enhancing medical reasoning and evidence-based decision-making.

\section{Related Work}

Systematic reviewing is a time‑consuming process that has recently attracted interest for automation (\citealp{9356351}, \citealp{han2024automating}). However, it has been observed that automation still faces limitations, with human reviewers outperforming current automated methods on such tasks (\citealp{yuan2022automate}).
The main goal of our study is to reflect the challenge of critical appraisal of medical articles, providing a benchmark for evaluating how well models can support both retrieval and critical reasoning in biomedical contexts.

To our knowledge, no existing dataset specifically targets the task of critical appraisal and analysis of scientific articles. However, similar resources for the general evaluation of biomedical NLP models are available: for instance, PubMedQA (\citealp{jin2019pubmedqadatasetbiomedicalresearch}) is an English-language dataset where questions are derived from the abstracts of biomedical research articles; MedQA (\citealp{jin2020diseasedoespatienthave}) offers an open-domain QA benchmark in English, simplified Chinese, and traditional Chinese, based on medical textbooks; SciDQA (\citealp{singh2024scidqadeepreadingcomprehension}) provides a collection of questions grounded in full-length scientific articles with figures and images. 
However, none of these datasets explicitly focus on the critical evaluation of research methodology. They primarily assess factual comprehension or domain knowledge, rather than the ability to review a scientific article in terms of study design, methodology or limitations. In contrast, our dataset is specifically designed to capture and evaluate these skills through multiple-choice questions grounded in medical literature. 

While the French biomedical NLP landscape still lacks a dataset that aims to explore critical appraisal of scientific articles, other resources exist for adjacent tasks: MedFrenchmark (\citealp{quercia2024medfrenchmark}) and FrenchMedMCQA (\citealp{labrak2023frenchmedmcqafrenchmultiplechoicequestion}) focus on MCQA in the medical domain, though without contextual grounding in articles; CAS (\citealp{grabar:hal-01937096}) supports information extraction tasks; and QUAERO (\citealp{neveol14quaero}) provides annotations for named entity recognition. More broadly, DrBenchmark (\citealp{labrak2024drbenchmarklargelanguageunderstanding}) consolidates 20 different biomedical tasks in French, offering a comprehensive evaluation suite for LLMs.

Alongside, recent efforts have proposed RAG frameworks tailored for tasks in the biomedical field. For example, Li et al. \citeyearpar{li2024benchmarkingretrievalaugmentedlargelanguage} provide a systematic evaluation of RAG-based approaches across various biomedical applications. Studies (\citealp{he2025retrievalaugmentedgenerationbiomedicinesurvey}) have shown that RAG systems significantly outperform standard LLMs on tasks such as information extraction and question answering. BiomedRAG (\citealp{li2024biomedragretrievalaugmentedlarge}) simplifies the integration of retrieved knowledge by incorporating relevant passages into LLM inputs, while BioRAG (\citealp{wang2024bioragragllmframeworkbiological}) combines a scientific corpora with domain-specific embedding and hierarchical knowledge structures to improve biological question reasoning. 
While our approach does not employ RAG, it complements this line of research by addressing the critical appraisal of scientific articles, an aspect that remains underexplored in existing resources. 

These resources reflect growing interest in evaluating LLMs in the biomedical domain, but they do not yet address the evaluation of critical reading skills, grounded in scientific literature, which our dataset aims to address.

\section{Dataset overview}

CareMedEval (\textbf{C}ritical \textbf{A}ppraisal and \textbf{RE}asoning \textbf{Med}ical \textbf{Eval}uation) is a French dataset focused on evaluating critical appraisal skills in the medical field for scientific articles, as practiced in French medical education. 

This dataset is composed of 534 questions taken from \textit{Lecture Critique d'Articles} exams (LCA, \textit{Critical appraisal of research articles}), taken by sixth-year medical students in France. During the LCA exams, which last three hours, students are asked to answer a series of multiple-choice questions grounded in critical reading of a given scientific article. 
Students are required to critically analyze and interpret scientific biomedical articles, most of which are clinical studies published in peer-reviewed journals. These articles include observational studies like case-control or interventional studies such as randomized clinical trials, and cover a broad range of medical specialties like epidemiology, biostatistics or public health. Students are expected to demonstrate critical reasoning skills by classifying study types and methodological approaches, understanding the implications for clinical practice, recognizing potential biases or study limitations, and evaluating statistical evidence supporting the study conclusions. A sample of our dataset showing an excerpt from a scientific article, the given instruction prompt with a corresponding question, and answer choices is shown in \autoref{fig:multimodataset}.

\subsection{Dataset collection}

\begin{table*}[h]
    \centering
    \resizebox{\textwidth}{!}{%
    \begin{tabular}{llll}
        \toprule
        \textbf{Label} & \textbf{Description} & \textbf{Skills required} & \textbf{Support}\\
        \midrule
        design         & Identification of study design            & Information retrieval & 105 \\
        statistics     & Understanding and interpretation of statistics & General knowledge, Information retrieval & 239\\
        methodology    & Knowledge of scientific methodology       & General conceptual understanding & 219 \\
        limitations    & Critical review of biases and limitations & Contextual reasoning & 132 \\
        applicability  & Clinical relevance and applicability      & Contextual reasoning & 115 \\
        \bottomrule
    \end{tabular}
    }
    \caption{Labels by type of reasoning involved in the critical appraisal of biomedical articles. Each question in the dataset was annotated with one or more labels reflecting the cognitive and analytical skills required to answer it.}
    \label{tab:reasoning_skills}
\end{table*}

\paragraph{Source data} This dataset is built from two main sources:

\begin{itemize}
    \item \textbf{Epreuves Classantes Nationales (ECN) website}\footnotemark[2], where we can find the official national LCA exams,
    \item \textbf{Collège National des Enseignants de Thérapeutique (CNET) website}\footnotemark[3], which publishes mock LCA exams reviewed and approved by an educational committee to ensure alignment with real exams.
\end{itemize}

\footnotetext[2]{\url{https://www.cng.sante.fr/candidats/internats/concours-medicaux/etudiants/epreuves-classantes-nationales-ecn}}
\footnotetext[3]{\url{https://therap.fr/lca/}}

The articles used in these exams are publicly available genuine scientific papers. We used these two websites as they provide official and mock exams, freely accessible and suitable for corpus construction. The ECN website offers real past exams, adding authenticity and relevance to our dataset. As for the CNET website, it features training exams with professional corrections and commentary, which is particularly valuable for understanding the reasoning behind correct answers. To our knowledge, there are no other free and easily accessible online sources offering similar LCA exam content.

\paragraph{Languages} The scientific articles are in English. The questions, answers, and justifications are in French. We also provide a preliminary English translation of the French parts of the dataset, generated using Gemini 2.5 Flash. 

\paragraph{Annotations and labels} Each question in the CareMedEval dataset was manually annotated with one or more labels, reflecting the cognitive and analytical skills required to answer it. These labels were created specifically for this dataset by a medical expert with a background in general practice, based on their professional expertise and reference textbooks. These labels were designed to capture meaningful distinctions in the types of reasoning and knowledge needed to answer questions correctly.

The labels are listed in \autoref{tab:reasoning_skills} and represent different cognitive dimensions: 

\begin{itemize}
    \item \textbf{Information retrieval:} locating relevant information or data points in the article.
    \item \textbf{General knowledge:} recalling factual or foundational medical information.  
    \item \textbf{General conceptual understanding:} reasoning about underlying principles or relationships between medical concepts.  
    \item \textbf{Contextual reasoning:} interpreting information in the specific context of the article, integrating multiple pieces of evidence to form a judgment.
\end{itemize}

Since questions often target multiple dimensions of skills, they can be assigned more than one label. These labels can help determine whether certain categories of questions are more challenging than others for models, depending on the types of skills they require.

For data sourced from the ECN website, we manually corrected the exam questions with the help of a general practitioner, as the official answer keys were not publicly available for these past exams.

For data from CNET website, we also collected the correct answers and justifications provided by medical professionnals for a subset of questions. These justifications explain why some answers are correct or false, offering valuable insight into clinical reasoning and critical appraisal.

\subsection{Dataset structure}
CareMedEval is composed of 534 questions in JSON format and 37 articles available in PDF. 
Statistics of this dataset are shown in \autoref{tab:dataset_stats}. On average, each question contains 15.6 tokens and has 2.60 correct answers. Most questions have multiple correct answers: about 29\% have two or three correct options, 20\% have four, 19\% only have one, and a small portion (around 3\%) have five correct answers.

\begin{table}[h]
\centering
\footnotesize
\begin{tabular}{l r}
\toprule
\textbf{Statistic} & \textbf{Value} \\
\midrule
Questions (total) & 534 \\
With justifications & 204 \\
Vocabulary size & 1,273 words \\
Question length (avg.) & 15.6 tokens \\
Correct answers (avg.) & 2.6 \\
\midrule
Articles (total) & 37 \\
Questions per article (avg.) & 14.4 \\
\quad min / max & 8 / 16 \\
\midrule
Article length (avg.) & 5,675 tokens \\
\quad min / max & 2,747 / 8,332 \\
Article length (PDF, avg.) & 10 pages \\
\midrule
Abstract length (avg.) & 1,019 tokens \\
\quad min / max & 276 / 1,832 \\
\bottomrule
\end{tabular}
\caption{CareMedEval dataset statistics.}
\label{tab:dataset_stats}
\end{table}

Each question in the dataset is represented as a JSON object with the fields described in \autoref{table:fieldescriptions}.

Each question in the dataset is linked to a scientific article through the \texttt{id\_article} field, which serves as a reference key to the corresponding article file. There are on average 14.4 questions per article. The minimum is 8 and the maximum is 16. Articles are available in plain text (.txt) format for easier processing, and the original PDF versions are also included. The plain text files were generated using the PyMuPDF library\footnotemark[4], which extracts raw textual content from each page, without figures. We manually reviewed and corrected the formatting issues in the resulting files to ensure readability and to preserve the original structure of the articles as much as possible. Using the Byte-Pair Encoding tokenizer from the tiktoken\footnotemark[5] library, we found that scientific articles in our dataset contain between 2,747 and 8,332 tokens, with an average length of 5,675 tokens per article. For the abstracts alone, also included in our dataset and matched with their corresponding article IDs, the number of tokens ranges from 276 to 1,832, with an average length of 1,019 tokens.\\

\footnotetext[4]{\url{https://github.com/pymupdf/PyMuPDF}}
\footnotetext[5]{\url{https://github.com/openai/tiktoken}}

\begin{table*}[ht]
  \centering
  \footnotesize
  \begin{tabular}{ll}
    \toprule
    \textbf{Field} & \textbf{Description} \\ 
    \midrule
    \texttt{id} & A unique identifier for the question \\
    \texttt{id\_article} & Internal article ID \\
    \texttt{source\_exam} & URL of the exam or online resource \\
    \texttt{date\_exam} & Date of the exam \\
    \texttt{article\_link} & URL of the article \\
    \texttt{article\_date} & Article publication date \\
    \texttt{question} & Question as presented in the exam \\
    \texttt{answers} & Dictionary mapping each option label (A to E) to its full answer text \\
    \texttt{correct\_answers} & List of correct answers labels \\
    \texttt{essential\_answers} & List of essential answers for LCA grading (23)\\
    \texttt{unacceptable\_answers} & List of inadmissible answers for LCA grading (19)\\
    \texttt{labels} & Labels describing skills or knowledge required (see \autoref{tab:reasoning_skills}) \\
    \texttt{justification} & Expert-written explanation of correct and incorrect answers (204) \\
    \texttt{nb\_correct\_answers} & Number of correct answer options for the question \\
    \bottomrule
  \end{tabular}
  \caption{Field descriptions for the CareMedEval dataset. Each entry corresponds to a multiple-choice question associated with a biomedical research article. Some fields are specific to certain subsets and are designed to support fine-grained evaluation, reasoning analysis, and expert-based interpretation. Subset size is specified in parenthesis.
  }
  \label{table:fieldescriptions}
\end{table*}

For the PDF versions, articles are on average approximately 10 pages long, with around 5,400 words and 36,000 characters. On average, each article includes about 3.3 figures, for an estimated total of 123 figures across the entire dataset (approximations computed using the PyMuPDF library). The publication years range from 2007 to 2023. In addition of the plain text and PDF versions, each question includes a direct \texttt{article\_link} field pointing to the online HTML version of the article.

\section{Benchmark}

To systematically assess the critical appraisal abilities of LLMs, we introduce a dedicated benchmark built upon our dataset. It provides a structured evaluation framework that integrates multiple metrics, controlled scenarios, and a representative set of models to enable comprehensive and reproducible analysis.

\subsection{Metrics}

Our benchmark relies on four different metrics chosen to reflect the multiple-choice question-answer nature of the dataset: 
\begin{itemize}
    \item \textbf{Exact Match Ratio} (EMR) measures the proportion of questions for which the predicted set of answers exactly matches the gold standard. 
    \item \textbf{F1-score} is the harmonic mean of precision and recall, computed between predicted and gold answer sets. 
    \item \textbf{Hamming score} evaluates the proportion of correctly predicted labels (answer options) over the total number of possible labels, averaged over all questions. 
    \item \textbf{LCA score} is a custom metric inspired by the grading system used in the original LCA exam from which our dataset is derived. It reflects exam-style grading: for each question, a perfect match yields 1 point, one mismatch 0.5, two mismatches 0.25, and more than two mismatches or no response 0. In addiction, there are two other constraints to the LCA grading system: if a required (essential) answer is missing, the score is automatically 0, regardless of other matches. If an unacceptable answer (one that should never be selected) is included, the score is 0. The final LCA score is averaged over all questions.
\end{itemize}

In France, the LCA exam is part of the "\textit{Epreuves de Connaissances}" ("Knowledge Exams"), which account for 60\% of the final score in the national medical competition. Results for the LCA exam and for these knowledge exams in general are not publicly available. However, we know that students must achieve a minimum score of 14/20 (70\%) to advance to the next stage of the competition\footnotemark[6], which can be used as a reference point for assessing models' success on the task. 

\footnotetext[6]{\url{https://www.cng.sante.fr/epreuves-dematerialisees-nationales-edn}}

These metrics allow us to distinguish between full correctness (EMR), partial correctness (F1, Hamming), and real-world grading fairness (LCA score).

\subsection{Evaluation scenarios and prompts}

\begin{figure*}[t]
\centering
  \includegraphics[width=1\textwidth]{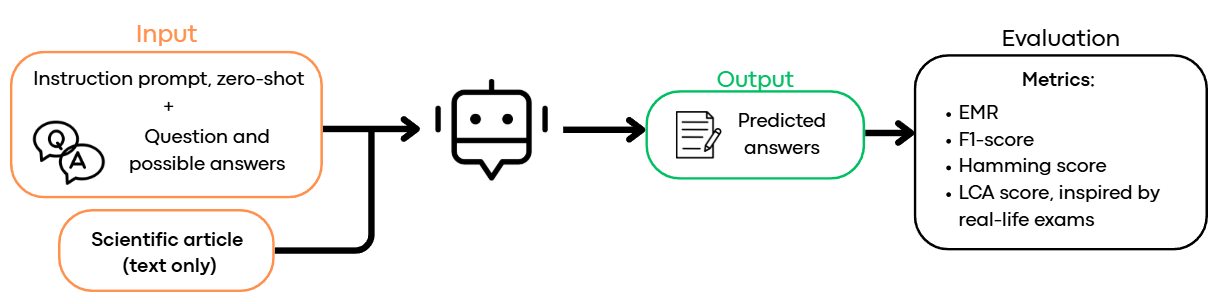}
  \caption{Overview of the model evaluation pipeline of the CareMedEval benchmark. The input consists of a zero-shot instruction prompt containing a question and possible answer choices, along with article (plain text only in our experiment setting). The model generates predicted answers, which are then evaluated using a set of quantitative metrics to assess performance.}
  \label{fig:evalpipeline}
\end{figure*}

We designed multiple evaluation scenarios in which the model was given an instruction-style prompt in French specifying the expected output format, a question, and the answer choices. Depending on the setting, the full article was provided, or only the abstract was included, or no article content was given. This pipeline is shown in \autoref{fig:evalpipeline}.

Each prompt was built by inserting the chosen context between a fixed prefix and suffix, containing instructions, the question and the answer choices (see appendix \ref{sec:evalprompt} for full prompt). Rather than dynamically truncating the article based on token limits, we predefined the context length in each scenario to ensure it remained within the acceptable range for the models being tested. Regardless of a model's maximum input size, all models were at least evaluated in two minimal context settings: with no article provided, and with the article abstract only. This approach allowed us to assess the impact of different levels of contextual information while avoiding any risk of exceeding model input limits.

We used a role-oriented instruction prompt, framing the model as a medical professional. While the impact of such role framing remains debated \cite{zheng2024ahelpfulassistantreally}, we kept it consistent across evaluation scenarios.

\subsection{Models}
To enable a comprehensive evaluation on this dataset, we selected models of varying sizes (from 8B to 120B parameters), architectures, domain specialization (general-purpose vs. biomedical-tuned), reasoning token generation capabilities, allowing us to assess the effect of the benchmark on different dimensions: Qwen3-8B/32B \cite{yang2025qwen3technicalreport}, II-Medical-8B \cite{intelligent_internet_ii_medical_8B}, Gemma3-27B-text-IT \cite{google_medgemma_27b_text_it}, MedGemma-27B-text-IT \cite{google_medgemma_27b_text_it}.  We also evaluated GPT-4.1 \cite{openai2024gpt4technicalreport}, GPT-4o-mini \cite{openai2024gpt4technicalreport} and GPT-OSS-20B/120B \cite{openai2025gptoss120bgptoss20bmodel}  to assess how frontier and derived models perform compared to smaller or domain-specific alternatives.

\section{Results and analysis}

We conducted a series of experiments on the CareMedEval dataset to evaluate the critical appraisal skills of a diverse pool of LLMs. While this dataset can support multiple experimental setups, we focus here on a MCQA task based solely on the textual content of the articles.

All experiments were run on an cluster equipped with NVIDIA L40-48GB and A100-80GB GPUs. Inference was performed using the vLLM \cite{kwon2023efficientmemorymanagementlarge} and Ollama\footnotemark[7] engines in float16 precision. The models relied on their native HuggingFace tokenizers (AutoTokenizer) for prompt encoding and truncation. Generation was fully deterministic (temperature = 0.0, top-p = 1.0), with a maximum of 8,000 generated tokens per prompt. Prompts were written in French and constrained the model to output only the letter(s) corresponding to the correct multiple-choice answers. The maximum prompt length was set to 31,000 tokens, with dynamic truncation applied exclusively to the article content.\\

\footnotetext[7]{\url{https://ollama.com/}}
 
In this section, we showcase some of the findings, while the complete set of results is available in the \href{https://github.com/bonzid/CareMedEval}{accompanying material}. Overall, GPT-4.1 demonstrates the highest performance across all evaluation scenarios, with Qwen3-32B consistently ranking as the second best. Notably, only four models surpass an EMR of 0.25 and none surpasses a LCA score of 0.70, which would be the minimal mark for the exam, highlighting the difficulty of the task. For comparison, GPT-4.1 reaches an EMR of 0.79 on the FrenchMedMCQA dataset \cite{labrak2023frenchmedmcqafrenchmultiplechoicequestion}. The difficulty of French medical MCQs from licensing examinations was highlighted in a previous study \cite{Alfertshofer2023}, where the "multiple correct answers" format appeared to be a factor contributing to models' poor performance on the task. This observation is consistent with our analysis of the subset presented in \ref{sec:wcnccontext}: models tend to perform better when the question explicitly requires a single correct answer. Interestingly, models specialized for the biomedical domain do not consistently outperform generalist models, exhibiting comparable performance at best.

\begin{table*}[h]
\centering
\begin{tabular}{lcccc}
\toprule
\multirow{2}{*}{\textbf{Model}} 
& \multicolumn{4}{c}{} \\
& EMR & F1 & Hamming & LCA \\
\midrule
Random (1-3 options) & 0.03 & 0.39 & 0.29 & 0.16 \\
Only first-option & 0.00 & 0.20 & 0.11 & 0.00 \\
Most 2 frequent answers & 0.03 & 0.45 & 0.33 & 0.18 \\
Most 3 frequent answers & 0.03 & 0.55 & 0.42 & 0.20 \\
\midrule
GPT4.1 & \textbf{0.49} & \textbf{0.84} & \textbf{0.78} & \textbf{0.68} \\
GPT4o-mini & 0.25 & 0.75 & 0.65 & 0.50 \\
    
\midrule
Qwen2.5-3B-Instruct & 0.10 & 0.59 & 0.46 & 0.31 \\
Qwen2.5-3B-GRPO-medical-reasoning & 0.11 & 0.57 & 0.45 & 0.30 \\
Qwen3-8B & 0.19 & 0.68 & 0.57 & 0.42 \\
II-Medical-8B & 0.13 & 0.51 & 0.43 & 0.30 \\
Qwen3-32B & \underline{0.37} & \underline{0.78} & \underline{0.70} & \underline{0.58} \\

\midrule
Gemma3-27B-text-IT & 0.27 & 0.75 & 0.65 & 0.51 \\
Medgemma-27B-text-IT & 0.28 & 0.73 & 0.63 & 0.50 \\

\bottomrule
\end{tabular}
\caption{Results of long-context models on the biomedical MCQA task of our dataset: zero-shot with full-text article provided. Bold values indicate the best results per metric, and underlined values the second-best. Baselines include random or frequency-based answer selection strategies.}
\label{32kcontext}
\end{table*}

\subsection{Comparison of models performance}

\subsubsection{Generalist vs. specialized}

The goal is to assess whether biomedical-pretrained or fine-tuned models 
offer a concrete advantage on medical tasks compared to the models they have been specialized from, under our hypothesis that specialized models are expected to perform better on a domain-specific critical reasoning task.  Results presented in ~\autoref{32kcontext} show comparable EMR scores between some specialized and generalist models (\textit{e.g}., Medgemma vs. Gemma3). In several cases, generalist models even outperform their specialized counterparts, such as Qwen3-8B vs. II-Medical-8B. 

To evaluate whether EMR differences between generalist and specialized models were significant, we applied McNemar's test, which compares paired outcomes by counting where one model is correct and the other is not. Despite noticeable differences in raw performance, the test shows that, in most cases, these gaps are not statistically significant (p $\ge$ 0.05). 
Significant differences were only observed for Qwen3-8B and II-Medical-8B. Consequently, our benchmark does not provide sufficient evidence to confirm our hypothesis. However, in line with prior works \cite{labrak2024drbenchmarklargelanguageunderstanding, dorfner2024biomedicallargelanguagesmodels}, our results are consistent with the observation that generalist models can perform competitively against domain-specialized models in medical tasks.

\subsubsection{Influence of article access on performance} \label{sec:wcnccontext}

\begin{figure}[htb]
  \includegraphics[width=\columnwidth]{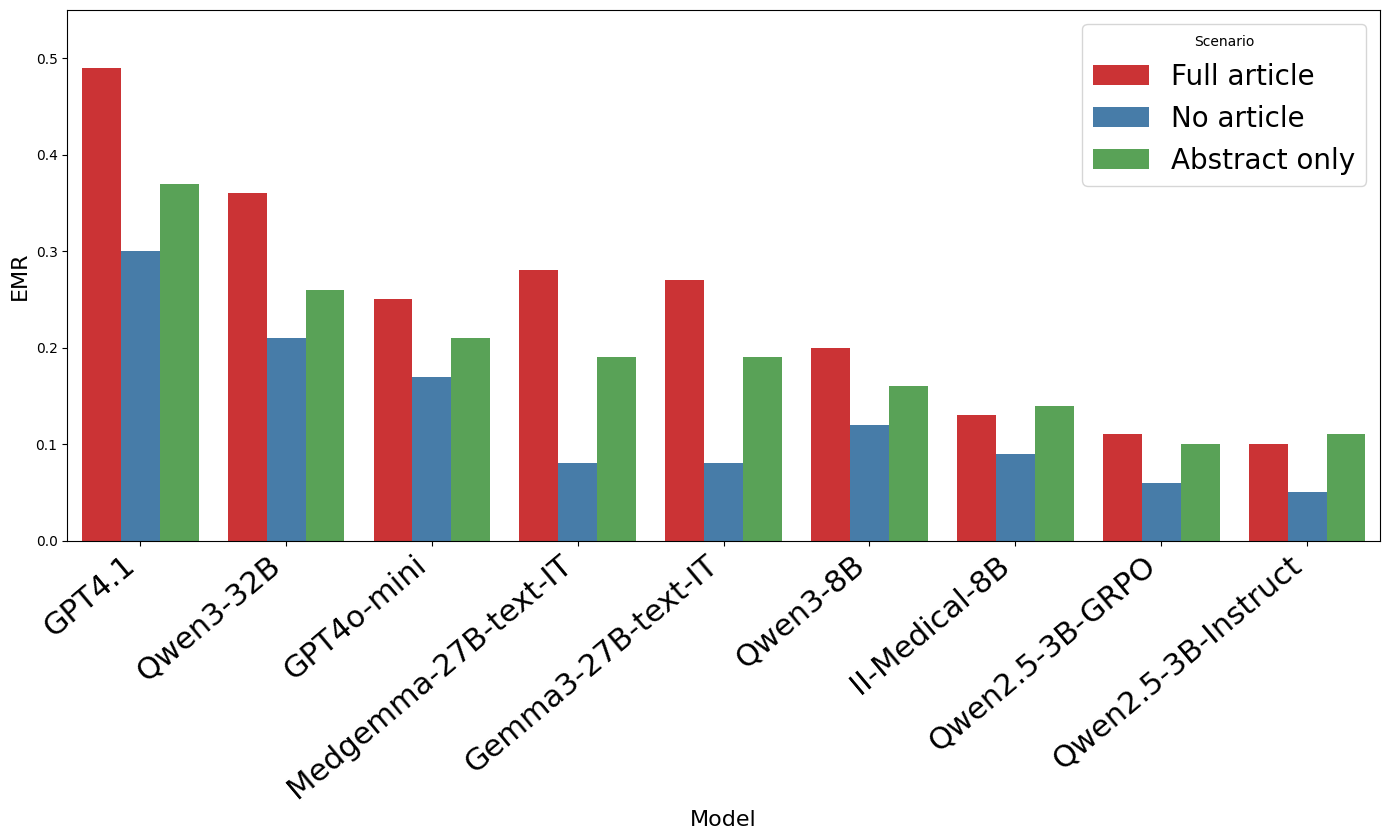}
  \caption{Exact Match Ratio comparison across different evaluation scenarios, illustrating model performance when provided with the full article, only the abstract, or no context (only the question and answer options with the instruction prompt).}
  \label{fig:emrscenario}
\end{figure}

To assess the importance of context for critical appraisal, we evaluated models under three different settings as  shown in ~\autoref{fig:emrscenario}. In addition to the question and answer choices, models were provided with either the full article, only the abstract, or no contextual information. The vast majority of questions in our dataset require access to the article to be answered correctly. While a few questions can be answered without reading the article, relying mostly on general medical knowledge, we still expect a performance drop when the article is not provided.

Performance varied significantly depending on the amount of context available. When given access to the full article, models achieved their highest scores; for example, GPT-4.1 reached an EMR of 0.49, and Qwen3-32B 0.36. While this confirms that access to the entire article improves question comprehension and accuracy, the overall performance remains moderate, reflecting the inherent difficulty of the task.  some
Moreover, it is possible that some articles may have been included, either fully or partially, in the models' pre-training data. In such cases, model predictions may rely on memorized content or pattern recall rather than genuine reasoning over the provided context. This form of data leakage could artificially inflate performance scores, particularly in the full-article setting, and should therefore be considered when interpreting the results.

With only the abstract, model performance dropped slightly but remained better than in the no-context condition. 
GPT-4o-mini showed slightly lower results with just the abstract compared to the full article, yet its performance remained competitive, suggesting that the abstract alone contains a substantial portion of the relevant information. Without any contextual input, all models experienced a notable decrease in performance, with EMR scores dropping by 5 to 15 points.

These findings highlight the importance of full-text access to maximize model performance on the task. While models like GPT-4.1 and Qwen3-32B demonstrate relative robustness even with limited or no context, by possibly mastering questions which do not require context, smaller models generally struggle to compensate for missing information.

We annotated a subset of 16 questions to indicate whether they require context to be answered correctly (field \texttt{requires\_context}, can be \texttt{true} or \texttt{false}). These annotations were based on general trends observed in model performance: whether models could answer questions correctly with or without access to the article.

Thanks to this subset, we can see some regularities and performance patterns with context or no context provided: some questions, particularly those directly related to specific aspects of the study (like \textit{"This is a study of:"}) consistently require the article for all models. Other questions show little difference in performance between the with- and without-context settings, 
suggesting that context is not necessary for these. Conversely, some questions (like \textit{"What was the main reason the data and safety monitoring board re-evaluated the sample size during the trial?"}) were answered more accurately without the article, suggesting the presence of a bias for this type of question rather than a benefit from using contextual information. 

\begin{figure}[t]
  \includegraphics[width=\columnwidth]{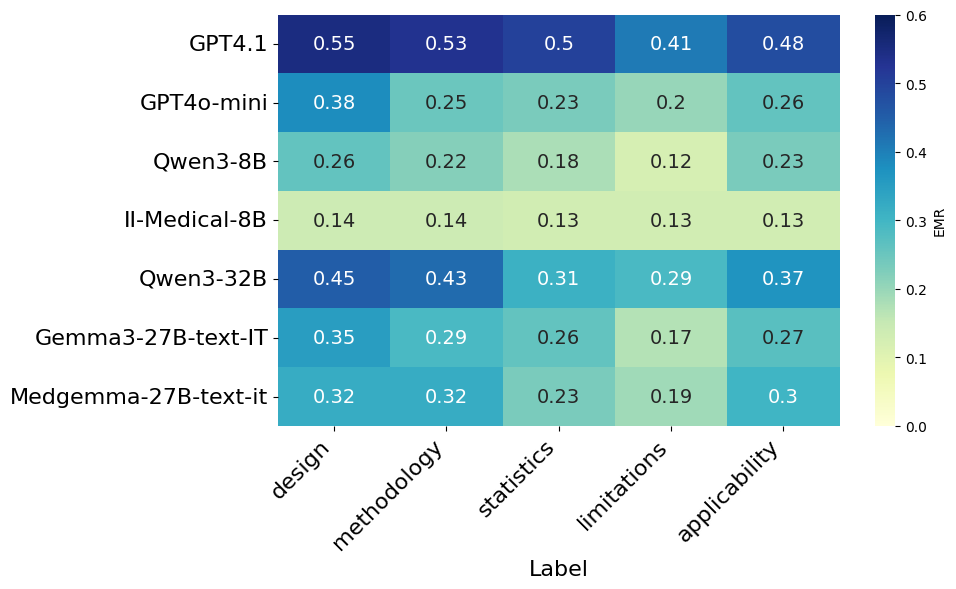}
  \caption{Heatmap of Exact Match Ratio by model and label for the MCQA task, illustrating performance differences across reasoning categories in the critical appraisal of scientific articles. Labels correspond to distinct cognitive skills required to answer the questions as described in ~\autoref{tab:reasoning_skills}.}
  \label{fig:labelheatmap}
\end{figure}

\begin{table*}[t]
\centering
\begin{tabular}{lcccc}
\toprule
\multirow{2}{*}{\textbf{Model}} 
& \multicolumn{4}{c}{} \\
& EMR & F1 & Hamming & LCA \\
\midrule
Qwen3-8B (w/o$\to$with) & 0.19$\to$0.35 & 0.68$\to$0.75 & 0.57$\to$0.66 & 0.42$\to$0.55 \\
Qwen3-32B (w/o$\to$with) & 0.37$\to$0.45 & 0.78$\to$0.81 & 0.70$\to$0.73 & 0.58$\to$0.64 \\
\midrule
GPT-OSS-20b (low$\to$high) & 0.36$\to$0.49 & 0.77$\to$0.81 & 0.60$\to$0.75 & 0.57$\to$0.66 \\
GPT-OSS-120b (low$\to$high)  & 0.46$\to$0.54 & 0.81$\to$0.85 & 0.74$\to$0.79 & 0.65$\to$0.71 \\
GPT4.1 (w/o$\to$with) & 0.49$\to$0.53 &  0.84$\to$0.85 & 0.78$\to$0.79 &  0.68$\to$0.71 \\
\bottomrule
\end{tabular}
\caption{Comparison of model performance according to "reasoning" tokens generation level on the CareMedEval benchmark.}
\label{results_reasoning}
\end{table*}

\subsubsection{Evaluation details by labels}

Each question in our dataset was annotated with one or more labels highlighting the cognitive and analytical skills required to answer it. These labels allow us to analyze which types of questions are more or less challenging for the models. These results are presented in ~\autoref{fig:labelheatmap}.

Models appear to struggle the most with questions labeled \textit{limitations}, which involve reviewing the biases or limitations of the study. This label typically requires fine-grained contextual understanding and often goes beyond what is explicitly stated in the text.
The low scores suggest that models are subject to difficulty with implicit critical reasoning.

The \textit{statistics} label, which requires understanding and interpreting statistical results, also shows lower performance compared to other categories. 
This can be partly explained by limitations in quantitative reasoning, but also by the fact that articles are provided in plain text format, excluding figures where statistical information is often presented.

Questions labeled \textit{design} and \textit{methodology} are those on which the models perform best. The top-performing models reach strong scores in these categories 
which may reflect the models' ability to recognize study structure and generalize research concepts as typically presented in medical articles.

\subsubsection{Impact of reasoning tokens generation on performance}

We evaluated the impact of explicit reasoning tokens generation on performance by comparing standard predictions (\textit{without reasoning}) to tests where models were prompted or trained to produce a reasoning sequence before answering (\textit{with reasoning}). For models that do not expose a mode without reasoning, we compare low and high reasoning effort presets. The results are available in \autoref{results_reasoning}. For each experiment we generated a single reasoning trace based on default parameters; on average, 879 tokens were generated ranging from 36 to 20,019 tokens across all models.

For GPT-4.1, reasoning is explicitly requested in the prompt, whereas for the Qwen3 models, we extract the reasoning part naturally generated by the model (content between \texttt{<think}> tags). We also evaluate GPT-OSS models \cite{openai2025gptoss120bgptoss20bmodel}, according to low and high reasoning profiles. Even though we do not show results here, the medium reasoning profile yields results between those of the low and high profiles, generally closer to the later.

Incorporating intermediate reasoning steps improves performance over no/low reasoning across all metrics, sometimes substantially, suggesting that generating explicit reasoning tokens helps models produce more accurate answers. This result suggests that CareMedEval indeed requires some form reasoning and can evaluate efforts from LLM makers to address this aspect. We leave the manual evaluation of reasoning quality to future work as it is non trivial to match the reasoning trace output by models with the human-written justifications available in the dataset.

\section{Conclusion}

In this work, we introduced an original dataset designed for evaluation of critical appraisal of scientific articles in the medical domain, combining data collection and expert annotation. We assess the suitability of the dataset by computing the performance of a range of models.

Overall, our experiments show that larger models like GPT-4.1 and Qwen3-32B tend to perform better on the task and domain-specialized biomedical models do not reliably outperform generalist models, often showing similar levels of performance. However, using our LCA score based on real-life medical exams, we find that none of the tested models without reasoning achieve the passing score that human candidates typically reach.

Providing the full-text article as context considerably improves model performance compared to using only abstracts or no context at all. This underlines the necessity of access to complete scientific information for accurate question answering and critical reasoning. Moreover, allowing models to generate reasoning tokens improves performance, highlighting that reasoning is essential for producing more reliable and contextually grounded answers in a critical appraisal task. 

In future work, we plan on extending the benchmark to vision LLMs that can leverage the content of figures which is sometimes referenced in questions or necessary to produce correct answers. We would also like to create an evaluation framework of the reasoning traces produced by models, compared to the justifications provided by experts.

\section{Limitations \& Ethics statement}

\paragraph{Limitations}

Our study has several limitations that should be acknowledged.

First, our evaluation only used the textual content of the articles, without including figures, tables, or other materials found in the original PDFs. These parts often have important information that helps understand a study better, like how it was designed or its results. Multimodal models, which have access to the full article content including these visuals, might perform better especially for questions about statistics or critical appraisal of the study’s weaknesses. Integrating retrieval-augmented generation (RAG) to dynamically select the most relevant sections of each article based on question keywords could also help models focus on the most informative content and improve accuracy. Beyond RAG integration, we aim to explore how altering the input article can help assess a model's ability to question and reason over modified scientific content.

Second, we evaluated a limited number of models, with a focus on general-purpose LLMs rather than domain-specific ones. Their performances may not fully reflect the potential of specialized biomedical models. In addition, our experiments used a fixed prompt structure without exploring prompt engineering variations or model-specific adaptations. Exploring alternative prompting strategies, including dynamic or adaptive prompts, could lead to improved performance.

Third, the manual annotation of justifications was performed by a small number of annotators, which may introduce variability and bias in the dataset. Expanding the number and diversity of annotators would help increase the reliability of the ground-truth justifications.

Fourth, the scientific articles included in our dataset are available on the internet and may have been part of the training data of the evaluated LLMs. This raises the possibility that some answers could benefit from memorization rather than genuine reasoning, potentially inflating performance metrics.

Finally, the dataset itself is relatively small (534 questions) and focused on a specific educational context (French medical LCA exams). Future work could benefit from expanding the dataset in size and scope to cover a broader range of biomedical topics and question formats. 

\paragraph{Ethics statement}

The goal of this work is to evaluate natural language processing technologies in order to better understand their capabilities and limitations, in particular within the context of the EU AI Act regulatory framework. We stress the importance of building useful tools for humans and the society, that do not decrease expertise or agency, both of which are critical in the medical domain.

\section{Acknowledgements}

This work was financially supported by ANR MALADES (ANR-23-IAS1-0005). Experiments were conducted using HPC resources provided by the Laboratoire d'Informatique et Systèmes (LIS) in Marseille. We thank Guillaume Lomet, Elie Antoine and Irina Illina for helpful feedback on an earlier version of this work.

\section{Bibliographical References}\label{sec:reference}
\bibliographystyle{lrec2026-natbib}
\bibliography{lrec2026-example}

\newpage

\section{Appendix}

\subsection{Evaluation prompt} \label{sec:evalprompt}
\begin{figure}[htbp]
\centering
\footnotesize
\begin{tcolorbox}[
  colback=white,
  colframe=black,
  boxrule=0.5pt,
  left=6pt,
  right=6pt,
  top=6pt,
  bottom=6pt
]

Vous êtes un médecin capable d’interpréter rigoureusement les données d’études médicales.  
À partir de l’article donné, répondez à la question à choix multiples suivante.  
Indiquez uniquement la ou les lettres correspondant aux bonnes réponses parmi : A, B, C, D, E.

\vspace{0.5em}

Votre réponse doit respecter strictement le format suivant :  
une ou plusieurs lettres, séparées par des virgules (ex. : B, E).

\vspace{0.8em}

\textbf{Article :}  
{article}
\vspace{0.8em}

\textbf{Question :}  
{question}
\vspace{0.8em}

\textbf{Choix de réponses :}  
{choices}

\vspace{0.8em}
\textbf{Votre réponse :}

\end{tcolorbox}
\caption{Prompt used for the evaluation, in French.}
\end{figure}

\begin{figure}[htbp]
\centering
\footnotesize
\begin{tcolorbox}[
  colback=white,
  colframe=black,
  boxrule=0.5pt,
  left=6pt,
  right=6pt,
  top=6pt,
  bottom=6pt
]

You are a physician capable of rigorously interpreting data from medical studies.  
Based on the given article, answer the following multiple-choice question.  
Indicate only the letter(s) corresponding to the correct answer(s) among: A, B, C, D, E.

\vspace{0.5em}

Your answer must strictly follow this exact format:  
one or more letters, separated by commas (e.g.: B, E).

\vspace{0.8em}

\textbf{Article:}  
{article}

\vspace{0.8em}

\textbf{Question:}  
{question}

\vspace{0.8em}

\textbf{Answer choices:}  
{choices}

\vspace{0.8em}

\textbf{Your answer:}

\end{tcolorbox}
\caption{Translated prompt from French to English. The French version was used for the experiments.}
\end{figure}

\end{document}